\documentclass{article}

\usepackage[preprint]{neurips_2023}
\usepackage[utf8]{inputenc} 
\usepackage[T1]{fontenc}    
\usepackage[dvipsnames]{xcolor}
\usepackage[colorlinks=true,linkcolor=Purple,citecolor=Emerald]{hyperref}
\usepackage{url}            
\usepackage{booktabs}       
\usepackage{amsfonts}       
\usepackage{nicefrac}       

\usepackage{microtype}      
\usepackage{caption}
\usepackage{multirow}
\usepackage{wrapfig}
\usepackage{xspace}
\usepackage{tcolorbox}
\usepackage{graphicx}
\usepackage{fancyhdr,lipsum}
\usepackage[export]{adjustbox}

\tcbuselibrary{most}

\vspace{-4em}
\title{TinyLlama: An Open-Source Small Language Model}
\vspace{-5em}
\author{Peiyuan Zhang$^{*}$ 
$\quad$ Guangtao Zeng$^{*}$
$\quad$ Tianduo Wang
$\quad$ Wei Lu \\
StatNLP Research Group\\
Singapore University of Technology and Design \\
\texttt{\{peiyuan\_zhang, tianduo\_wang, luwei\}@sutd.edu.sg}\\
\texttt{guangtao\_zeng@mymail.sutd.edu.sg} \\
}

\begin{document}
\maketitle
\renewcommand{\thefootnote}{\fnsymbol{footnote}}
\footnotetext[1]{The first two authors contributed equally.}
\renewcommand{\thefootnote}{\arabic{footnote}}

\vspace{-2.5em}
\begin{figure}[h]
\centering
\includegraphics[width=0.45\linewidth]{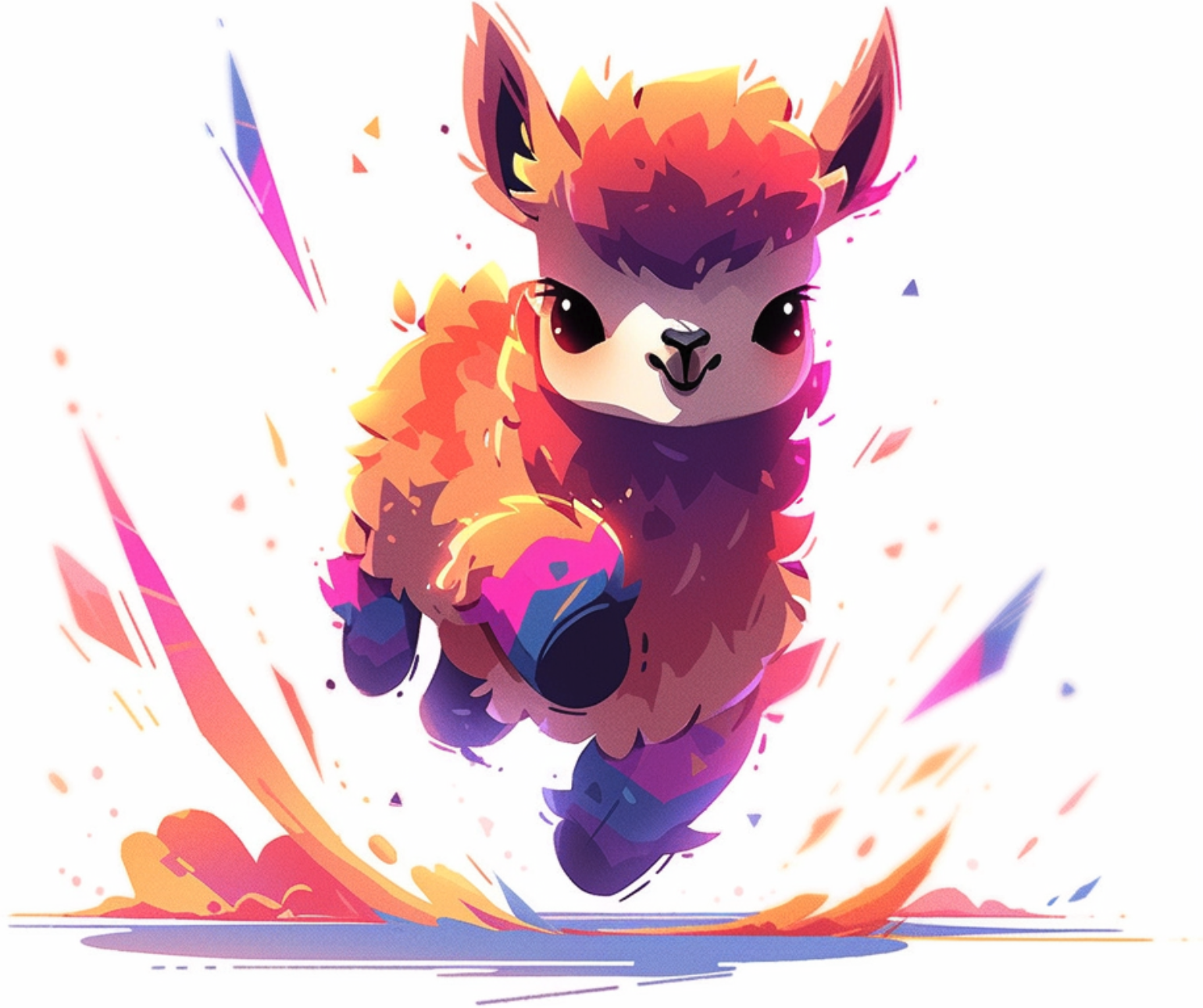}
\label{fig:your_label}
\end{figure}

\begin{abstract}
We present TinyLlama, a compact 1.1B language model pretrained on around 1 trillion tokens for up to 3 epochs\footnote{\textcolor{black}{Our latest model, TinyLlama v1.1, is only trained for 2 trillion tokens. More details about this latest version will be elaborated in the later section.}}. 
Building on the architecture and tokenizer of Llama 2~\citep{touvron2023llama2},  TinyLlama leverages various advances contributed by the open-source community, e.g., FlashAttention~\citep{dao2023fa2} and Lit-GPT~\citep{lit-gpt}, achieving better computational efficiency.
Despite its relatively small size, TinyLlama demonstrates remarkable performance in a series of downstream tasks.
It significantly outperforms existing open-source language models with comparable sizes.
Our model checkpoints and code are publicly available on GitHub at
\url{https://github.com/jzhang38/TinyLlama}.

\end{abstract}

\section{Introduction}

Recent progress in natural language processing (NLP) has been largely propelled by scaling up language model sizes~\citep{brown2020gpt3,chowdhery2022palm,touvron2023llama,touvron2023llama2}.
Large Language Models (LLMs) pre-trained on extensive text corpora have demonstrated their effectiveness on a wide range of tasks~\citep{gpt4,touvron2023llama2}.
Some empirical studies demonstrated emergent abilities in LLMs, abilities that may only manifest in models with a sufficiently large number of parameters, such as few-shot prompting~\citep{brown2020gpt3} and chain-of-thought reasoning~\citep{wei2022cot}.
Other studies focus on modeling the scaling behavior of LLMs~\citep{kaplan2020scaling,hoffmann2022chinchilla}.
\cite{hoffmann2022chinchilla} suggest that, to train a compute-optimal model, the size of the model and the amount of training data should be increased proportionally. 
This provides a guideline on how to optimally select the model size and allocate the amount of training data when the compute budget is fixed.

Although these works show a clear preference on large models, 
    the potential of training smaller models with larger datasets remains under-explored.
Instead of training compute-optimal language models, \cite{touvron2023llama} highlight the importance of the inference budget, instead of focusing solely on training compute-optimal language models.
Inference-optimal language models aim for optimal performance within specific inference constraints.
This is achieved by training models with more tokens than what is recommended by the scaling law ~\citep{hoffmann2022chinchilla}.
\cite{touvron2023llama} demonstrates that smaller models, when trained with more data, can match or even outperform their larger counterparts.
Also, \cite{thaddee2020cd} suggest that existing scaling laws~\citep{hoffmann2022chinchilla} may not predict accurately in situations where smaller models are trained for longer periods.

Motivated by these new findings, this work focuses on exploring the behavior of smaller models when trained with a significantly larger number of tokens than what is suggested by the scaling law~\citep{hoffmann2022chinchilla}.
Specifically, we train a Transformer decoder-only model~\citep{vaswani2017transformer} with 1.1B parameters using up to 3 trillion tokens.
To our knowledge, this is the first attempt to train a model with 1B parameters using such a large amount of data.
Following the same architecture and tokenizer as Llama 2~\citep{touvron2023llama2}, we name our model TinyLlama.
TinyLlama shows competitive performance compared to existing open-source language models of similar sizes.
Specifically, TinyLlama surpasses both OPT-1.3B~\citep{zhang2022opt} and Pythia-1.4B~\citep{biderman2023pythia} in various downstream tasks.
%
%
%
Our TinyLlama is open-source, aimed at improving accessibility for researchers in language model research. 
We believe its excellent performance and compact size make it an attractive platform for researchers and practitioners in language model research.

\vspace{-1em}
\section{Pre-training}
\vspace{-1em}
This section outlines the pre-training process of TinyLlama. 
First, We introduce how we composed the datasets used for pre-training. 
Next, we describe the model architecture and the hyperparameters implemented during the pre-training phase.
\vspace{-0.3em}
\subsection{Pre-training data}
\vspace{-0.2em}
\textcolor{black}{
%
We utilize a blend of natural language and code data to pre-train TinyLlama, drawing from two primary sources:
    SlimPajama~\citep{cerebras2023slimpajama} and the training data of StarCoder~\citep{li2023starcoder}.
A detailed introduction to these two datasets is provided below.
}
\vspace{-1em}
\paragraph{SlimPajama}

\textcolor{black}{
It is a high-quality corpus specifically created for training large language models.
This corpus, derived from RedPajama~\citep{together2023redpajama}, underwent additional cleaning and deduplication processes.
The original RedPajama corpus, an open-source research project, was designed to replicate the pretraining data of Llama~\citep{touvron2023llama} and contains over 1.2 trillion tokens.
After extensive filtering to remove low-quality and duplicated content, SlimPajama retains only 50\% of the original tokens from RedPajama.
}
\vspace{-1em}
\paragraph{StarCoder Training Dataset}

\textcolor{black}{
This dataset is collected to train StarCoder~\citep{li2023starcoder}, a powerful open-source large code language model. 
It contains code data in 86 programming languages.
In addition to code, it also includes GitHub issues and text-code pairs that involve natural languages.
As SlimPajama also contains data from GitHub\footnote{\url{https://huggingface.co/datasets/cerebras/SlimPajama-627B}},
    we remove the GitHub subset from SlimPajama and 
        only sample code-related data from the StarCoder training dataset to avoid duplication.
}

\textcolor{black}{
After merging these two datasets, we obtained a combined total of approximately 950 billion tokens for pre-training, 
    processed using the Llama tokenizer~\citep{touvron2023llama,touvron2023llama2}.
TinyLlama is then trained on these tokens across approximately three epochs, cumulatively processing 3 trillion tokens.
The training data is derived from the SlimPajama and Starcoder datasets at a sampling ratio of approximately 7:3.
}

\subsection{Architecture}
%

\textcolor{black}{
    Following the series of the Llama models~\citep{touvron2023llama,touvron2023llama2},
        our model is based on the decoder-only Transformer~\citep{vaswani2017transformer}.
    The details of hyperparameters for our model architecture are provided in Table~\ref{tab:model}.
}
\vspace{-10pt}
\begin{table}[htbp]
  \centering
  \normalsize
  \caption{ 
    The details of model architecture.
  }
  \scalebox{1.0}{
  \begin{tabular}{lcccccccc}
    \toprule
    \textbf{Hidden size} & \textbf{Intermediate Hidden Size} & \textbf{Context Len}& \textbf{Heads} & \textbf{Layers} & \textbf{Vocab size}\\

    \midrule    
    2,048      & 5,632  & 2,048       & 32    & 22  & 32,000  \\
    \bottomrule
  \end{tabular}
  }
  \label{tab:model}
\end{table}

\vspace{-1pt}
\paragraph{Positional embedding} 
We use Rotary Positional Embedding (RoPE)~\citep{su2021roformer} to inject positional information into our model.
RoPE is a widely adopted method recently used by many mainstream large language models, such as PaLM~\citep{anil2023palm}, Llama~\citep{touvron2023llama,touvron2023llama2}, and Qwen~\citep{bai2023qwen}.
\vspace{-1em}
\paragraph{Pre-norm and RMSNorm} 
\textcolor{black}{
Following Llama~\citep{touvron2023llama,touvron2023llama2},
    we adopt pre-norm, i.e., normalize the inputs of each sub-layer in Transformer, 
    instead of post-norm to stabilize the training process.
Additionally, the RMSNorm~\citep{DBLP:conf/nips/ZhangS19a} normalization function is applied to improve the training efficiency.
}
\vspace{-1em}
\paragraph{SwiGLU} 
\textcolor{black}{
Instead of using the traditional ReLU activation function,
    we follow Llama~\citep{touvron2023llama,touvron2023llama2} and adopt SwiGLU~\citep{DBLP:journals/corr/abs-2002-05202},
    i.e., the combination of the Swish activation function and the Gated Linear Units (GLU)~\citep{dauphin2017glu}.
}
\vspace{-1em}
\paragraph{Grouped-query Attention} 
To reduce memory bandwidth overhead and speed up inference, we use grouped-query attention~\citep{DBLP:journals/corr/abs-2305-13245} in our model.
We have 32 heads for query attention and use 4 groups of key-value heads.
With this technique, the model can share key and value representations across multiple heads without sacrificing much performance.
\vspace{-1em}
\subsection{Speed Optimization}
\vspace{-0.5em}
\paragraph{Fully Sharded Data Parallel (FSDP)}
During training, our codebase has integrated FSDP\footnote{\url{https://huggingface.co/docs/accelerate/usage_guides/fsdp}} to leverage multi-GPU and multi-node setups efficiently. This integration is crucial in scaling the training process across multiple computing nodes, which significantly improves the training speed and efficiency.
\vspace{-1em}
\paragraph{FlashAttention} 
Another critical improvement is the integration of FlashAttention-2~\citep{dao2023fa2}, an optimized attention mechanism.
The repository also provides fused layernorm, fused cross entropy loss, and fused rotary positional embedding, which together play a pivotal role in boosting computational throughput.
\vspace{-1em}
\paragraph{xFormers}
\textcolor{black}{
We have replaced the original SwiGLU module with the fused SwiGLU module from the xFormers \citep{xFormers2022} repository, further enhancing the efficiency of our codebase.
With these features, we can reduce the memory footprint, enabling larger batch size during 1.1B model training.
}

\begin{wraptable}{r}{0.40\textwidth}
  \vspace{-12pt}
  \centering
  \caption{ 
     Comparison of our training speed with Pythia-1.0B and MPT-1.3B.
  }
  \scalebox{1.0}{
  \begin{tabular}{lc}
    \toprule
     & \textbf{GPU Hours} \\
    \midrule    
    Pythia-1.0B      & 4,830     \\
    MPT-1.3B         & 7,920     \\
    TinyLlama        & 3,456     \\
    \bottomrule
  \end{tabular}
  }
  \label{tab:comparison}
  \vspace{-10pt}
\end{wraptable}
\vspace{-1em}
\paragraph{Speed Comparison with Existing Models} 
\textcolor{black}{
After implementing these speedup modules,
    we achieved a training throughput of 24,000 tokens per second per A100-40G GPU.
We compare our training speed with existing models, 
    Pythia-1.0B~\citep{biderman2023pythia} and MPT-1.3B\footnote{\url{https://huggingface.co/mosaicml/mpt-1b-redpajama-200b}},
    in terms of the GPU hours required to train 300 billion tokens, as detailed in Table~\ref{tab:comparison}.
The results indicate that our codebase significantly enhances training efficiency compared to these models.
}

\vspace{-0.5em}
\subsection{Training}
We build our framework based on lit-gpt~\citep{lit-gpt}.
In adhering to Llama 2~\citep{touvron2023llama2}, we employ an autoregressive language modeling objective during the pretraining phase.
Consistent with Llama 2's settings, we utilize the AdamW optimizer~\citep{DBLP:conf/iclr/LoshchilovH19}, setting $\beta_1$ at 0.9 and 
$\beta_2 $ at 0.95.
Additionally, we use a cosine learning rate schedule with a maximum learning rate of $4.0\times 10^{-4} $ and a minimum learning rate of $4.0 \times 10 ^ {-5} $.
We use 2,000 warmup steps to facilitate optimized learning.
We set the batch size as 2M tokens.
We assign weight decay as 0.1 and use a gradient clipping threshold of 1.0 to regulate the gradient value.
We pretrain TinyLlama with 16 A100-40G GPUs in our project.

\subsection{Version 1.1}

\begin{figure}[t]
\centering
\includegraphics[width=0.8\linewidth]{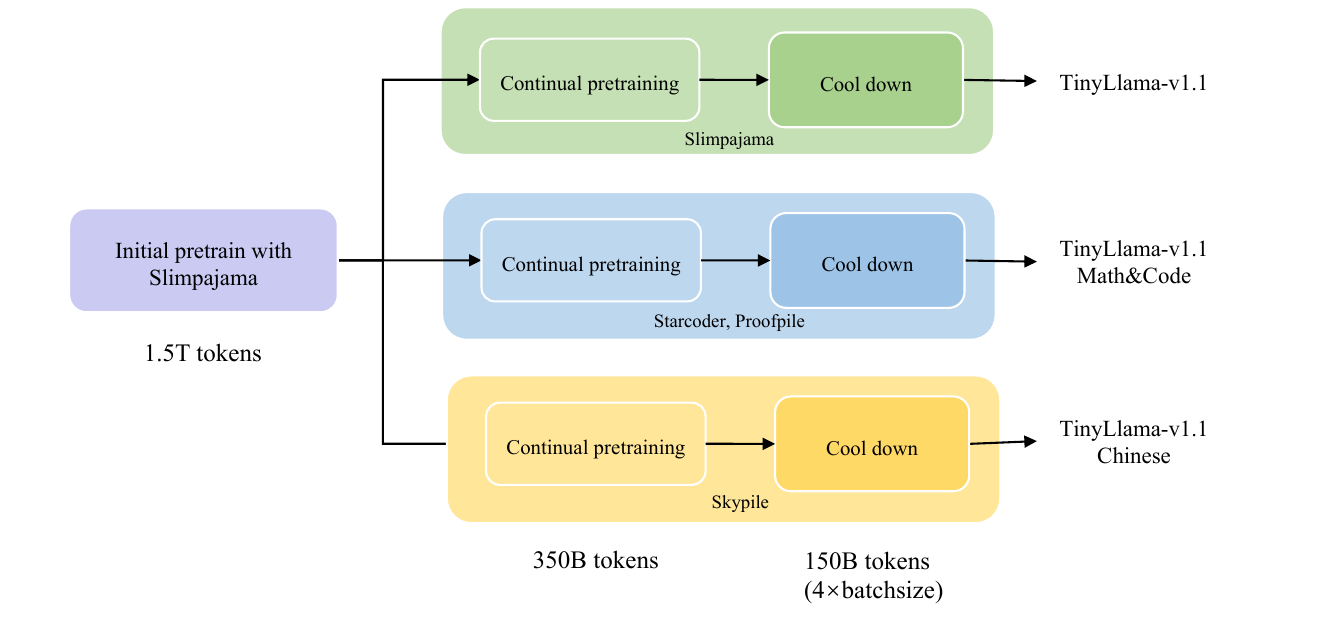}
\vspace{-8pt}
\caption{ 
   {The pre-training stages and specialization pipeline of TinyLlama v1.1.}
  }
\label{fig:overview}
\end{figure}

\textcolor{black}{
During the open, live phase of pre-training the first version of TinyLlama, a few implementation issues were identified within the training framework.
For example, there were a few issues related to the learning rate scheduler\footnote{\url{https://github.com/jzhang38/TinyLlama/issues/27}} 
and data loading processes\footnote{\url{https://github.com/jzhang38/TinyLlama/issues/67}}.
We therefore re-trained a new model from scratch after fixing those issues.
The new model is named \textbf{TinyLlama v1.1}.
In addition to fixing all those implementation issues, we took the opportunity to incorporate several key modifications in TinyLlama v1.1:}
\begin{itemize}
    \item To reduce communication overhead, we only shard the model parameters within a node in FSDP. \textcolor{black}{We trained with a cluster consisting of 16 nodes, each equipped with four A100-40G GPUs, and set the batch size to approximately 1.8 million tokens.}
    \item We reduced the total number of pre-training tokens from 3 trillion to 2 trillion. Despite the reduction in training data volume, a marginal improvement in performance on downstream tasks was observed, as compared to the original TinyLlama (refer to Section~\ref{sec:results}).
    \item Expanding beyond the singular pre-training phase of the original model, we introduced a three-stage pre-training process inspired by recent research \citep{wei2023skywork}. This includes basic pre-training, continual pre-training targeted at specific domains, and a cooldown phase. An illustrative overview of this approach is provided in Figure~\ref{fig:overview}, with detailed discussions to follow in subsequent sections.
\end{itemize}


\paragraph{Basic pre-training}
In the first stage of the pre-training, we trained our model with only SlimPajama~\citep{cerebras2023slimpajama} to develop its commonsense reasoning capabilities. 
We only trained 1.5 trillion tokens during this stage, setting the foundation for more specialized training in subsequent stages.

\paragraph{Continual pre-training with specific domain}
During this phase, we diversified the training by incorporating three distinct types of corpora. The first corpus, identical to the basic pre-training stage, solely consisted of SlimPajama data~\citep{cerebras2023slimpajama}. The second corpus combined code and mathematical content, leveraging integrations with Starcoder~\citep{li2023starcoder} and Proof Pile 2~\citep{azerbayev2023llemma}. 
\textcolor{black}{For Starcoder, we only considered the ``Python'' and ``Jupyter'' splits of the original Starcoder dataset.}
The third corpus focused on Chinese language data, utilizing Skypile~\citep{wei2023skywork}.  This strategic corpus diversity facilitated the development of three main TinyLlama v1.1 model variants, each tailored to a different need:
\begin{itemize}
\item \textbf{TinyLlama v1.1}: A foundational model for general applications.
\item \textbf{TinyLlama v1.1 - Math\&Code}: Enhanced specifically for mathematical and coding tasks.
\item \textbf{TinyLlama v1.1 - Chinese}: Specialized for processing and understanding Chinese text.
\end{itemize}
In this stage, all three variants are trained with 350 billion tokens.
For the Math\&Code and Chinese variants, we linearly increase the sampling proportion of the domain-specific corpus for the beginning 6 billion tokens.
This warmup sampling increasing strategy was designed to gradually adjust the pre-training data distribution, aiming to ensure smoother and more stable training.
Following this initial phase of adaptive sampling, we maintained a consistent sampling strategy for the remainder of the training until approximately 1.85 trillion tokens. Detailed of the data sampling ratios are provided in Appendix~\ref{appendix:data_sampling}.

\vspace{-1em}
\paragraph{Cooldown}

\begin{figure}[htbp]
\centering
\includegraphics[width=0.6\linewidth]{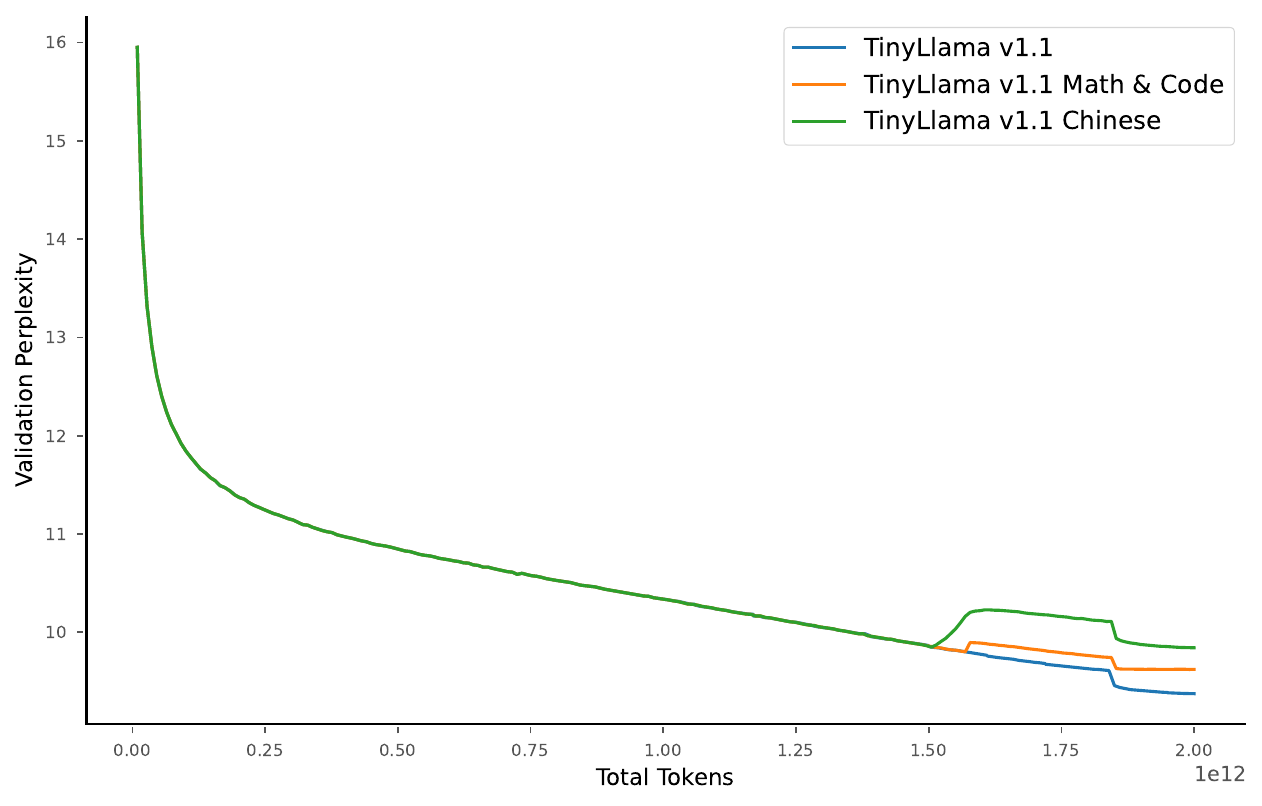}
\caption{ 
    Training Loss for TinyLlama v1.1 models.
  }
\label{fig:loss_curve}
\end{figure}

Implementing a cooldown phase is essential for enhancing model convergence towards the end of the pre-training process. 
Traditionally, this is achieved by modifying the learning rate, as seen in approaches like MiniCPM~\citep{hu2024minicpm} and DeepSeek LLMs~\citep{bi2024deepseek}. However, due to the use of a cosine schedule for our model, the learning rate is already low at the later stage of training.
Hence, we opted to modify the batch size instead. Specifically, during the cooldown stage, the batch size was increased from 1.8 million tokens to 7.2 million tokens, while maintaining the original cosine learning rate schedule. This adjustment was applied uniformly across all variants, with each undergoing an additional 150 billion tokens of training during this phase.
The training curves for all three variants are shown in Figure~\ref{fig:loss_curve}.
%
\vspace{-0.5em}
\vspace{-2em}
\section{Results}\label{sec:results}
\vspace{-1em}
We evaluate TinyLlama on a wide range of commonsense reasoning and problem-solving tasks and compare it with several existing open-source language models with similar model parameters.
\vspace{-1em}
\paragraph{Baseline models}
We primarily focus on language models with a decoder-only architecture, comprising approximately 1 billion parameters.
Specifically, we compare TinyLlama with OPT-1.3B~\citep{zhang2022opt}, Pythia-1.0B, and Pythia-1.4B~\citep{biderman2023pythia}.

\vspace{-0.5em}
\paragraph{Commonsense reasoning tasks} To understand the commonsense reasoning ability of TinyLlama,
we consider the following tasks: Hellaswag~\citep{zellers2019hellaswag}, 
OpenBookQA~\citep{mihaylov2018obqa}, 
WinoGrande~\citep{sakaguchi2021winogrande}, 
ARC-Easy and ARC-Challenge~\citep{clark2018arc}, 
BoolQ~\citep{clark2019boolq}, 
and PIQA~\citep{bisk2020piqa}.
We adopt the Language Model Evaluation Harness framework~\citep{evalharness} to evaluate the models.
Following previous practice~\citep{biderman2023pythia}, the models are evaluated in a zero-shot setting on these tasks.
The results are presented in Table~\ref{tab:pretrained_res}.
We notice that all TinyLlama models outperform baselines on many of the tasks and obtain the highest averaged scores.
As for the 3 TinyLlama v1.1 models, we found that the model can achieve the best hellaswag performance while only training on Slimpajama.
It is interesting to observe that for the Chinese model, even though the continual pretraining stage involves 50\% Chinese data, the model still maintains good performance in the English benchmark.

\begin{table}[htbp]
  \centering
  \scriptsize
  \caption{ 
    Zero-shot performance on commonsense reasoning tasks.
  }
  \begin{tabular}{lcccccccc}
    \toprule
    & \textbf{HellaSwag}	& \textbf{Obqa} &	\textbf{WinoGrande}	& \textbf{ARC-c}	& \textbf{ARC-e}	& \textbf{boolq}	& \textbf{piqa} & \textbf{Avg} \\

    \midrule    
    OPT-1.3B\xspace      &  53.65	&33.40	& 59.59&	29.44	&50.80	&60.83	&72.36	& 51.44\\
    Pythia-1.0B\xspace      &  47.16	&31.40	&53.43&	27.05	&48.99	&57.83	&69.21	&48.30\\
    Pythia-1.4B\xspace      &  52.01	&33.20	&57.38&	28.50	& 54.00	&\textbf{63.27}	&70.95	&51.33\\
    \midrule
    TinyLlama v1.0\xspace      &  59.20	 & 36.00&	59.12&	30.12 & 55.25&	57.83	& 73.29	& 52.99  \\
    TinyLlama v1.1\xspace      &  \textbf{61.47}	&\textbf{36.80}&	59.43&	32.68&\textbf{55.47}&	55.99	&\textbf{73.56}	& 53.63  \\
    TinyLlama v1.1 Math\&code\xspace    &  60.80 &  36.40 &  \textbf{60.22} &  \textbf{33.87} & 55.20 & 57.09 & 72.69 &  \textbf{53.75} \\
    TinyLlama v1.1 Chinese\xspace     &  58.23 &  35.20 &  59.27 &  31.40 & 55.35 & 61.41 & 73.01 &  53.41 \\
    
    \bottomrule
  \end{tabular}
  \label{tab:pretrained_res}
  \vspace{-0.5em}
\end{table}
\vspace{-0.5em}
\paragraph{Problem-solving tasks}
We also evaluate TinyLlama's problem-solving capabilities using the InstructEval benchmark \citep{DBLP:journals/corr/abs-2306-04757}. This benchmark includes the following tasks:

\begin{itemize}
\item Massive Multitask Language Understanding (MMLU) \citep{hendrycks2020mmlu}: This task is used to measure a model's world knowledge and problem-solving capabilities across various subjects. We evaluate the models in a 5-shot setting.
\item BIG-Bench Hard (BBH) \citep{suzgun2022challenging}: This is a subset of 23 challenging tasks from the BIG-Bench benchmark \citep{srivastava2022beyond} designed to measure a language model's abilities in complex instruction following. The models are evaluated in a 3-shot setting.
\item Discrete Reasoning Over Paragraphs (DROP) \citep{dua2019drop}: This reading comprehension task measures a model's math reasoning abilities. We evaluate the models in a 3-shot setting.
\item HumanEval \citep{DBLP:conf/kdd/ZhengXZDWXSW0LS23}: This task is used to measure a model's programming capabilities. The models are evaluated in a zero-shot setting.
\end{itemize}

The evaluation results are presented in Table~\ref{tab:pretrained_Instrust}.
We observe that TinyLlama demonstrates better problem-solving skills compared to existing models.
\textcolor{black}{Moreover, after pretraining with a corpus containing more code and math data}, TinyLlama v1.1 Math\&Code can quickly obtain relevant ability and have a significant improvement in HumanEval and Drop compared to the other models, even though this model did not consume lots of code and math corpus at first 1.5T tokens.

\vspace{-1em}
\begin{table}[htbp!]
  \centering
  \scriptsize
  \caption{ Performance of problem-solving tasks on the InstructEval Benchmark.
  }
  \begin{tabular}{lccccc}
    \toprule
     & \textbf{MMLU}	&\textbf{BBH}&	\textbf{HumanEval}&	\textbf{DROP} & \textbf{Avg.} \\
     & 5-shot& 3-shot& 0-shot& 3-shot& \\

    \midrule    
    OPT-1.3B\xspace      & 24.90 &	28.57 &	{\color{white} 0}0.00 &	14.32 & 16.95\\
    Pythia-1.0B\xspace      & 25.70 &	28.19 &	{\color{white} 0}1.83 &	{\color{white} 0}4.25 & 14.99\\
    Pythia-1.4B\xspace      & 25.41 &	29.01 &	{\color{white} 0}4.27  &	12.27 & 17.72 \\
    \midrule
        TinyLlama v1.0\xspace      & 25.34 &29.65 &	{\color{white} 0}9.15	& 15.34 & {19.87} \\

        TinyLlama v1.1\xspace      & \textbf{26.58} & 29.27 & {\color{white} 0}6.71 & 15.21 & 19.44 \\
        TinyLlama v1.1 Math\&Code \xspace      & 24.60 & 26.35 & \textbf{15.24} & \textbf{18.54} & \textbf{21.18} \\
        TinyLlama v1.1 Chinese \xspace      & 26.35 & \textbf{30.04} & {\color{white} 0}4.88 & 15.75 & 19.26 \\
    \bottomrule
  \end{tabular}
  \label{tab:pretrained_Instrust}
\end{table}

\vspace{-1em}

\paragraph{Evaluation for Chinese tasks}
To evaluate the Chinese understanding and reasoning ability of TinyLlama v1.1, we test our models with the following {\color{black}tasks}: xwinograd~\citep{emelin-sennrich-2021-wino}, xstorycloze~\citep{lin-etal-2022-shot}, xnli~\citep{conneau-etal-2018-xnli}, {\color{black}and} xcopa~\citep{ponti-etal-2020-xcopa}.
They are all multilingual multiple choice question answering tasks but we only consider Chinese in this {\color{black}evaluation exercise}.
We summarize the overall performance across a suite of benchmarks in Table~\ref{tab:chinese}.
\textcolor{black}{After continual pretraining on the Chinese corpus, we observe that TinyLlama v1.1 Chinese outperforms other models in most tasks.
Surprisingly, we also find that TinyLlama v1.0 and TinyLlama v1.1 Math\&Code can achieve a much better performance than the original TinyLlama v1.1.
We initially hypothesize that involving more code data helps improving the multilingual abilities.
Upon checking the starcoder and Proof Pile datasets, we discovered that approximately 3.5\% of the Python corpus and 4\% of the Jupyter corpus consist of Chinese text.
This may explain the improvement in the Chinese benchmark performance for TinyLlama v1.0 and TinyLlama v1.1 Math\&Code.
}

\begin{table}[htbp]
  \centering
  \scriptsize
  \caption{ 
    Zero-shot performance on Chinese understanding tasks.
  }
  \begin{tabular}{lccccc}
    \toprule
    & \textbf{xwinograd\_zh}	& \textbf{xstorycloze\_zh} &	\textbf{xnli\_zh}	& \textbf{xcopa\_zh} & \textbf{Avg} \\

    \midrule    
    OPT-1.3B\xspace      &  54.96 &  48.38 &  33.45 &  53.00 &  47.45 \\
    Pythia-1.0B\xspace      &  60.31 &  48.97 &  33.77 &  55.20 &  49.56\\
    Pythia-1.4B\xspace     &  60.51 &  50.56 &  35.38 &  53.00 &  49.86 \\
    \midrule
    TinyLlama v1.0\xspace   &  68.45 &  54.53 &  33.69 &  56.80 &  53.37  \\
    TinyLlama v1.1\xspace   &  55.75 &  47.98 &  \textbf{34.70} &  51.40 &  47.46  \\
    TinyLlama v1.1 Math\&Code\xspace    &  63.69 &  56.32 &  33.57 &  57.00 &  52.64  \\
    TinyLlama v1.1 Chinese\xspace     &  \textbf{74.20} &  \textbf{60.29} &  33.77 &  \textbf{65.20}&  \textbf{58.37} \\
    
    \bottomrule
  \end{tabular}
  \label{tab:chinese}
\end{table}

\section{Conclusion}

In this paper, we introduce TinyLlama, an open-source, small-scale language model.
With its compact architecture and promising performance, TinyLlama can enable end-user applications on mobile devices, and serve as a lightweight platform for testing a wide range of innovative ideas related to language models. We also additionally verify the effectiveness of multi-stage training and data schedule through our v1.1 series. To promote transparency in the open-source language model pre-training community, we have released all relevant information, including our pre-training code, intermediate model checkpoints, and the details of our data processing steps. We hope that TinyLlama will make a valuable contribution to the community as we aim to democratize language model research and provide useful insights for future research in this field.

\section*{Acknowledgements}
We express our gratitude to the open-source community for their strong support during the open, live phase of our research. Special thanks go to Qian Liu, Longxu Dou, Hai Leong Chieu, and Larry Law for their help to our project.
\textcolor{black}{We would like to thank the National Supercomputing Centre (NSCC) Singapore for their support in training our model v1.1.}
%
This research/project is supported by Ministry of Education, Singapore, under its Academic Research Fund (AcRF) Tier 2 Programme (MOE AcRF Tier 2 Award No.: MOE-T2EP20122-0011), Ministry of Education, Singapore, under its Tier 3 Programme (The Award No.: MOET320200004), the National Research Foundation Singapore and DSO National Laboratories under the AI Singapore Program (AISG Award No: AISG2-RP-2020-016), an AI Singapore PhD Scholarship (AISG Award No: AISG2-PhD-2021-08-007), an SUTD Kick-Starter Project (SKI 2021\_03\_11), and the grant RS-INSUR-00027-E0901-S00. Any opinions, findings and conclusions or recommendations expressed in this material are those of the authors and do not reflect the views of the funding agencies.

\bibliographystyle{apalike}
\bibliography{custom}

\appendix



Based on this observation, we reduced the number of tokens to 2T in total in the second run.
Instead of training a single base model, we first pretrained the model for 1.5T tokens and then continually pretrained it with different types of corpora. This resulted in three variants, TinyLlama v1.1, TinyLlama v1.1 Math\&Code and TinyLlama v1.1 Chinese.



\section{Data sampling ratio for TinyLlama v1.1}\label{appendix:data_sampling}
The data sampling ratio for TinyLlama v1.1 Math\&Code and Chinese model are shown in Table~\ref{tab:math_code_sampling} and~\ref{tab:chinese_sampling}.

\begin{table}[htbp]
  \centering
  \small
  \caption{ 
    Data sampling ratio for TinyLlama-Math\&Code
  }
  \begin{tabular}{lccc}
    \toprule
     & \textbf{Basic pretraining}	& \textbf{Continual pretraining with specific domain}	& \textbf{Cooldown}	\\

    \midrule    
    \textbf{Slimpajama} \xspace    & 100.0\%  &  75.0\%	& 75.0\% \\
    \textbf{Starcoder}\xspace    & -  &   	15.0\%	& 	15.0\% \\
    \textbf{Proof pile}\xspace    & -     &  	10.0\%	& 	10.0\% \\
    \bottomrule
  \end{tabular}
  \label{tab:math_code_sampling}
\end{table}

\begin{table}[htbp]
  \centering
  \small
  \caption{ 
    Data sampling ratio for TinyLlama-v1.1-Chinese
  }
  \begin{tabular}{lccc}
    \toprule
     & \textbf{Basic pretraining}	& \textbf{Continual pretraining with specific domain}	& \textbf{Cooldown}	\\

    \midrule    
    \textbf{Slimpajama} \xspace    & 100.0\%  &  50.0\%	& 50.0\% \\
    \textbf{Skypile}\xspace    & -  &   		50.0\%	& 		50.0\% \\
    \bottomrule
  \end{tabular}
  \label{tab:chinese_sampling}
\end{table}

\end{document}